\documentclass[11pt]{article}

\usepackage{fullpage}
\usepackage{sidecap}

\usepackage{graphicx}
\usepackage{times}

\begin{document}

\title{Studying Collective Human Decision Making and Creativity with
  Evolutionary Computation}

\author{Hiroki Sayama \thanks{Collective Dynamics of Complex Systems
    Research Group / Departments of Bioengineering \& Systems Science
    and Industrial Engineering, Binghamton University, Binghamton, New
    York 13902-6000. sayama@binghamton.edu} \and Shelley D. Dionne
  \thanks{Center for Leadership Studies, School of Management,
    Binghamton University, Binghamton, New York
    13902-6000. sdionne@binghamton.edu}}

\date{}

\maketitle

\begin{abstract}
We report a summary of our interdisciplinary research project
``Evolutionary Perspective on Collective Decision Making'' that was
conducted through close collaboration between computational,
organizational and social scientists at Binghamton University. We
redefined collective human decision making and creativity as evolution
of ecologies of ideas, where populations of ideas evolve via continual
applications of evolutionary operators such as reproduction,
recombination, mutation, selection, and migration of ideas, each
conducted by participating humans. Based on this evolutionary
perspective, we generated hypotheses about collective human decision
making using agent-based computer simulations. The hypotheses were
then tested through several experiments with real human
subjects. Throughout this project, we utilized evolutionary
computation (EC) in non-traditional ways---(1) as a theoretical
framework for reinterpreting the dynamics of idea generation and
selection, (2) as a computational simulation model of collective human
decision making processes, and (3) as a research tool for collecting
high-resolution experimental data of actual collaborative design and
decision making from human subjects. We believe our work demonstrates
untapped potential of EC for interdisciplinary research involving
human and social dynamics. \footnote{This paper is an extended version
  of \cite{ieeealife2013}.}
\end{abstract}


\section{Introduction}

In Artificial Life and other computational science and engineering
research areas, evolutionary computation (EC) has been widely used as
metaheuristics for nonlinear search, design and optimization
\cite{bentley1999,bentley2002}. From a different point of view,
however, EC also has great potential as a theoretical model that
operationalizes complex evolutionary processes in a simple, tractable
formalism. One application area of EC as a modeling tool is collective
human decision making and creativity
\cite{klein2003,kerr2004,salas2004}, which typically involves
high-dimensional nonlinear problem space, nontrivial societal
structure, within-individual cognitive and behavioral patterns, and/or
between-individual diversity.

In our project ``Evolutionary Perspective on Collective Decision
Making'' \cite{website}, we used EC as a concise, yet sufficiently
comprehensive, theoretical/computational model of the dynamics of
collective human decision making. We proposed a fundamental shift of
the viewpoint from the properties and behaviors of human participants
(as in most existing literature) to the properties and behaviors of
ideas being discussed. Collective decision making and creative
processes are redefined as evolution of ecologies of ideas over a
social network habitat, where populations of ideas evolve via
continual applications of evolutionary operators such as reproduction,
recombination, mutation, selection, and migration of ideas, each
conducted by participating humans. This paper summarizes how we
utilized EC in several non-traditional ways in this project.

\section{Theoretical Framework}
\label{theory}

The framework we propose views idea generation and selection in
collective human decision making as an evolutionary process, by
shifting the viewpoint from individual participants' personal
properties and decisions to the dynamical evolution of ideas being
discussed. Obviously, this is a drastic simplification of real humans'
social group dynamics that are very complex, which could involve
various interests, motives and strategies that would hugely influence
discussion outcomes. Here, we limit ourselves to studying how much we
could model, understand and predict group decision making by focusing
on temporal changes of ideas and describing them as evolutionary
processes.

In our framework, several key concepts developed and used in
evolutionary theory can be mapped to collective decision making as in
Table \ref{table1}. Many evolutionary operators are conceivable as a
representation of diverse human behaviors, as described below.

\begin{table}[t]
\centering
\caption{Mapping of concepts from evolutionary theory to concepts in
  human decision making (items with asterisks (*) do not exist in real
  biology, however).}
{\small
\begin{tabular}{lll}
\hline
Category & Concepts in evolution & Concepts in human decision making \\
\hline
Genetics & Genetic possibility space & Problem space \\
& Genome & Idea (a set of choices for all aspects of the problem) \\
& Locus on a genome & Aspect of the problem \\
& Allele & Specific choice made for an aspect \\
\hline
Ecology & Population & A set of ideas being discussed \\
& Environment & A group of individual participants and their utility functions \\
& Fitness & Utility value of an idea (either perceived or real) \\
& Adaptation & Increase of utility values achieved by the idea population \\
\hline
Evolutionary & Replication & Advocacy (increase of relative popularity of an idea) \\
Operator & Mutation & Minor modification of an idea \\
& Recombination & Production of a new idea by crossing multiple ideas \\
& Subtractive selection & Criticism (narrowing of diversity of ideas based on their fitness) \\
& Intelligent point mutation* & Improvement of an existing idea \\
& Random generation* & ``Crazy'' inspiration out of nowhere \\
\hline
\end{tabular}
}
\label{table1}
\end{table}

For selection-oriented operators, replication of an idea is a form of
positive selection, which represents advocacy of a particular idea
under discussion. Similarly, criticism against an idea may be
considered a form of negative, subtractive selection that eliminates
ideas with poor utility from the population. Subtractive selection
serves to reduce the number of ideas under consideration. Both
positive and negative selections seek to narrow decision possibilities
based on ``fitness'' among decision makers, in the expectation that it
will increase the overall fitness of ideas being discussed.

For variation-oriented operators, random point mutation adds a copy of
an idea with point mutations, making random changes to existing ideas
by asking ``what if''-type questions. Likewise, intelligent point
mutation initially begins like random point mutation, however several
variations from the original idea may be tried internally within a
human individual, and then the idea with the highest perceived fitness
will be selected and proposed. Intelligent point mutation is not
present in biological evolution but still is relevant to our
framework, because it represents hill climbing nature of human
thinking. Recombination and random generation also may be considered
within variation. Recombination represents the creation of a new idea
by crossing two (or more) existing ideas. Random generation represents
a sudden inspiration of a completely novel idea unrelated to the
existing ideas (which may arise entirely randomly or by drawing on
aspects of the individual's personal experiences outside the bounds of
the discussion). These variation operations enhance the exploratory
capabilities of the population, but generally reduce their immediate
fitness.

Here we note that the specific choices of evolutionary operators
presented in this paper are neither a focus nor a main contribution of
our work. Rather, our emphasis is on the significance of the proposed
modeling framework itself, with which one can develop many different
evolutionary operators based on the human behavioral patterns of
interest and easily implement them in simulations.

We are not the first to consider idea dynamics instead of participant
dynamics in collective decision making. For example, classic research
on the Delphi methodology \cite{turoff} investigated the dynamics of
idea convergence through standardized voting mechanisms that experts
can use as a tool to assess ideas and decisions. More relevant to our
framework is group decision support research via computer mediated
communication (CMC) \cite{connolly,dennis,vitharana}. Evolutionary
theory, in contrast, will offer a more generic yet mechanistic
framework to model the whole processes from idea generation to
decision selection, by which one can understand and produce hypotheses
about how the specifics of decision frameworks, such as bias, team
structure, and homogeneity/heterogeneity of individual and group
decision criteria, may impact decision processes and outcomes. Our
framework is also related to the concept of memes \cite{dawkins} and
memetic evolutionary studies \cite{aunger} that attempt to describe
human cultural evolution through the transfer of pieces of cultural
information over human populations. Our research is distinct from
memetics in that we specially focus on collective decision making that
takes place in timescale orders of magnitude shorter than that of
human cultural evolution. Moreover, our research aims to produce
prescriptive implications for general or specific tendencies of
collective decision making dynamics to help improve real-world human
decision making problems, which may also contribute to memetics that
is so far mostly descriptive in nature.

\section{Computer Simulation Model}
\label{sim}

We applied the evolutionary framework introduced above to develop a
computer simulation model of group decision making processes
within a small-sized, fully connected social network structure
\cite{iccs2007}. This is a straightforward application of EC to the
evolution of ideas, with additional assumptions made to represent
multiple human participants with different worldviews (i.e., fitness
evaluation criteria) and different behavioral traits (i.e., relative
frequencies of evolutionary operator usage).

\begin{figure}[t]
\centering
\includegraphics[width=0.9\columnwidth]{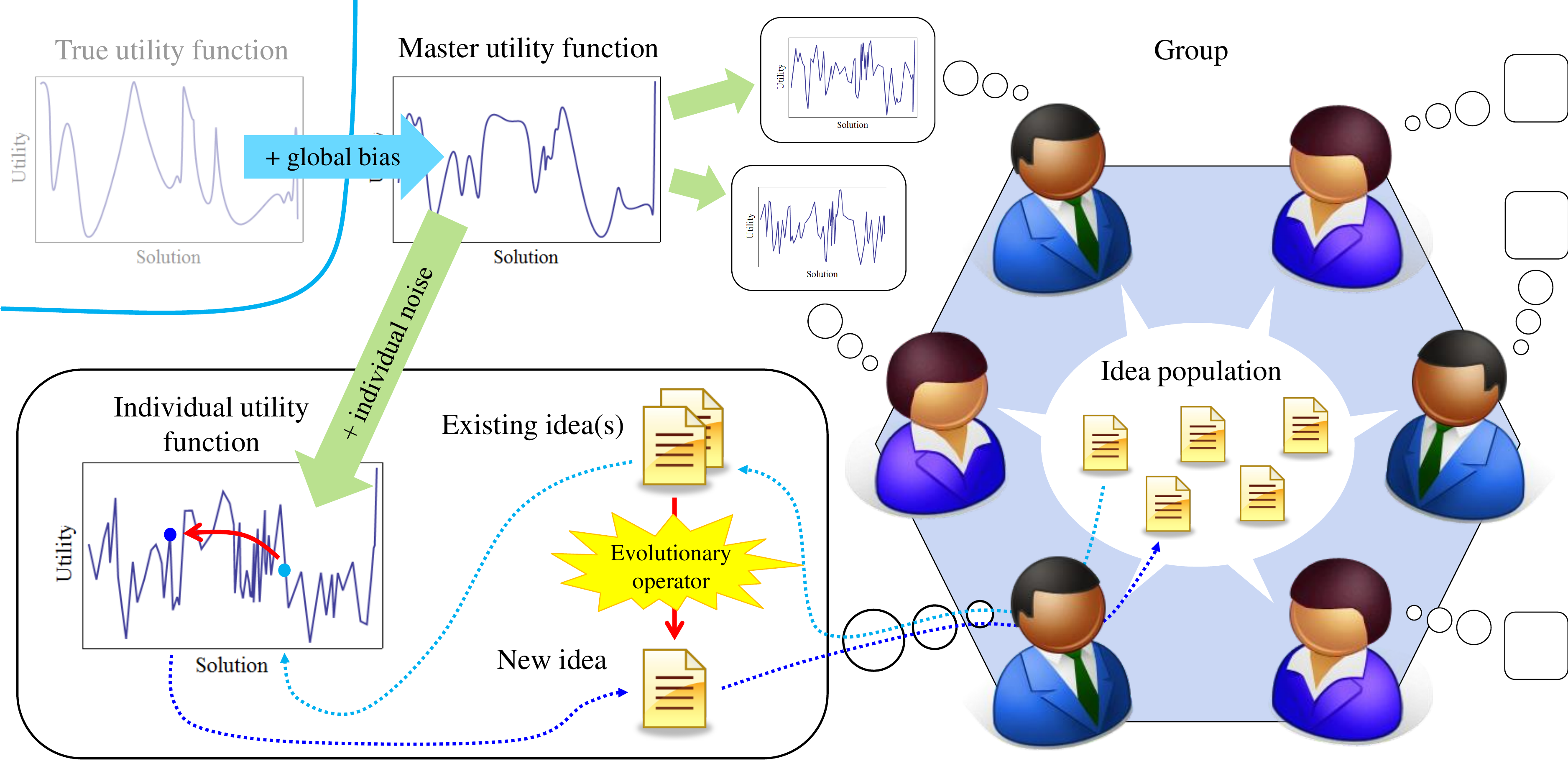}
\caption{An overview of our computer simulation model of group
  decision making processes.}
\label{comp-model}
\end{figure}

Our model consists of $N$ agents that are initiated with a shared
population of $k$ randomly generated ideas and then take turns in
applying various evolutionary operators to the idea population for $T$
iterations (Fig. \ref{comp-model}). Individuals always act in the same
order and groups always demonstrate a full rotation. The total number
of actions performed on the population in a simulation is thus
$NT$. In the population of ideas being discussed, there may exist
multiple copies of an identical idea, which represents the relative
popularity of that idea among group members. We assumed that actions
were performed on single copies, not the equivalence class of all idea
replicates.

The group of agents is situated in an $M$-dimensional binary problem
space, with $2^M$ possible ideas. For a simulation, every idea has a
utility value specified by a master utility function $U$ that is
unavailable to group members. Individuals perceive idea utility values
based on their own utility functions $U_j$ constructed by adding noise
to $U$. A semi-continuous assignment of utility values is defined over
the problem space in the following way. First, $n$ representative
ideas $S = \{v_i | i = 1 \ldots n\}$ are randomly generated, where
each $v_i$ represents one idea made of $M$ bits. One idea is assigned
the maximum fitness value, 1, and another, the minimum fitness value,
0, to ensure the range of utility values are always $[0,1]$. The
remaining $n-2$ ideas are assigned a random real value between 0 and
1. The master utility function is defined over its whole domain by
interpolation using ideas in $S$ and their utility values, i.e.,
\begin{equation}
U(v) = \frac{\sum_{i=1}^n U(v_i) D(v_i,v)^{-2}}{\sum_{i=1}^n D(v_i,v)^{-2}} ,
\end{equation}
where $v \not\in S$ is the idea in question, $U(v_i)$ is the utility
value of a representative idea $v_i \in S$, and $D(v_i,v)$ is the
Hamming distance between $v_i$ and $v$. By using the weighted average,
the master utility function is defined to have a moderately ``smooth''
surface in a high-dimensional problem space. Such smoothness of the
utility function gives room for intelligent group decision making to
outperform merely random decision making. 

Each individual agent in a group will have a different set of utility
values for the possible ideas. Individual utility functions $U_j(v)$
($j=1 \ldots N$) are generated by adding random white noise to the
master utility function so that
\begin{equation}
U_j(v) \in [\max (U(v) - \nu, 0), \min (U(v) + \nu, 1)]
\end{equation}
for all $v$, where $\nu$ is the parameter that determines the
amplitude of noise. An example of such an individual utility function
in contrast to the master utility function can be found in
Supplemental Materials (Fig.~S1). Since the
direction of noise added to individual utility function varies
stochastically from agent to agent, the parameter $\nu$ represents the
amount of {\em within-group heterogeneity}. The agents do not have
access to others' utility functions.
 
In addition to individual deviations from the master utility function,
we also investigated the effect of {\em group-level bias}, i.e.,
common deviations of the master utility function from the ``true''
utility function. To simulate this, we introduce a new step in the
generation of individual utility functions, in which the master
utility function $U$ differs from the true utility function
$U_T$. Specifically, a bias $\beta$ is imposed on the true utility
function both by flipping bits with probability $0.25\beta$ per bit on
representative ideas in $S$, and by adding a random noise in $[-\beta,
  \beta]$ to utility values of the representative ideas. Their utility
values are then renormalized to the range [0, 1]. The master utility
function is generated from these biased representative idea
set. Subsequent methods follow as described above. Group-level bias
represents fidelity of information at the group level, where $\beta=0$
denotes perfect understanding of the problem (at least as a
collective), and complete lack of understanding is asymptotically
approached as bias increases. Note that having group-level bias does
not influence the dynamics of group discussion processes; it only
affects the true utility value of the final group decision.

Regarding the agent behavior in discussion, we identified the
following six evolutionary operators available to the agents. Although
not an exhaustive list, these six operators reflect common forms of
action in evolution (either biological or informational) and can be
used as analogy for actions in collective human decision making. Some
operators use a preferential search algorithm to stochastically search
the idea population, where $r_p$ ideas are randomly selected and
ranked according to their individual utility values by each individual
agent, and then the best or worst idea is selected, depending on the
nature of the operator.
\begin{description}
\item[Replication] adds an exact copy of an idea selected from the
  population of ideas back onto the population. The idea is chosen for
  replication with the above-mentioned preferential search
  algorithm. This represents an advocacy of a particular idea under
  discussion.
\item[Random point mutation] adds a copy of an idea with point
  mutations, flipping of bits at each aspect of a problem with a
  probability $p_m$. The idea on which the operator acts is chosen
  from the population with the preferential search algorithm. This
  represents an attempt of making random changes to the existing
  ideas, reflected in asking ``what if''-type questions.
\item[Intelligent point mutation] selects an idea from the population
  with a preferential search algorithm, makes several ($r_m$)
  offspring of the idea by adding random point mutations, and selects
  that of the highest individual utility for addition to the
  population. This represents a proposal of an improved idea derived
  from existing ideas under discussion.
\item[Recombination] chooses one idea at random and one with a
  preferential search algorithm. It then creates two offspring from
  the two parent ideas with multiple point crossover: parent ideas are
  aligned by aspects, for each of which there is a probability $p_s$
  of switching their contents. Then the better offspring is added back
  to the population. This represents a creation of a new idea by
  crossing multiple existing ideas.
\item[Subtractive selection] deletes from the population the idea with
  the worst individual utility identified by the (negative)
  preferential search algorithm. This represents a criticism against a
  bad idea.
\item[Random generation] adds a randomly generated idea to the
  population. This represents a sudden inspiration of a completely
  novel idea that is unrelated to the existing ideas under discussion.
\end{description}
Relative frequencies of the use of those operators are assigned to
each agent as its own behavioral tendency, which is one of the
experimental parameters being varied. Parameter values used for the
simulations (provided in Supplemental Materials; Table
S1) were set so as to be reasonable in view of typical
real group decision making settings. We tested several minor
variations for each parameter value, confirming that the results were
not qualitatively different from the ones presented below. As
experimental conditions, the within-group heterogeneity $\nu$ and the
group-level bias $\beta$ were varied from 0 to 1.2 by increments of
0.2.

Once a simulation is completed, the following two metrics of group
performance are measured on the resulting idea population: True
utility of the most supported idea (to characterize the quality of the
decision made by the group), and entropy remaining in the idea
population (to characterize the level of consensus within the
group). The latter measurement is based on the classical information
theoretic definition
\begin{equation}
H = -\sum_{i=1}^l p(x_i) \log_2 p(x_i) ,
\end{equation}
where $l$ is the number of different types of ideas in the population,
and $p(x_i)$ is the ratio of the number of the $i$-th type of idea to
the entire population of ideas. The maximal possible entropy for a
group playing a sufficient number of rounds is $M$, which occurs when
there are exactly $2^M$ distinct ideas, all equally represented. Since
the entropy represents how many more bits would be needed to
completely specify the single final group decision, it can be assumed
that $M - H$ is a quantitative measure that intuitively means the
effective number of aspects of the problem on which the group has
formed a cohesive opinion. To rescale this to the range between 0 and
1, we use $(M - H) / M$ as a measurement of the level of convergence
of group decision.

\begin{figure}[t]
\centering
\begin{tabular}{ll}
(a) & (b) \\
\includegraphics[width=.49\columnwidth]{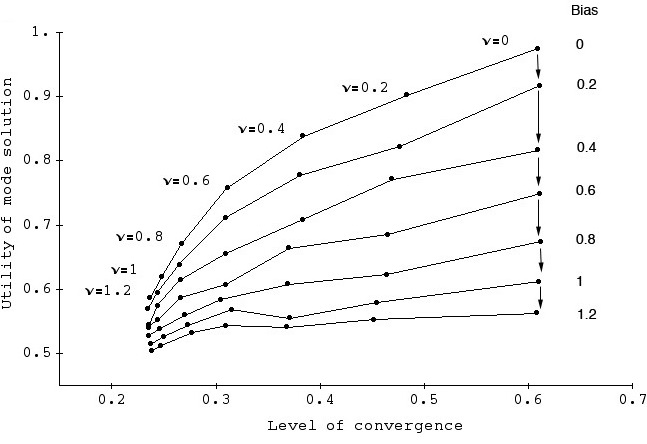} &
\includegraphics[width=.49\columnwidth]{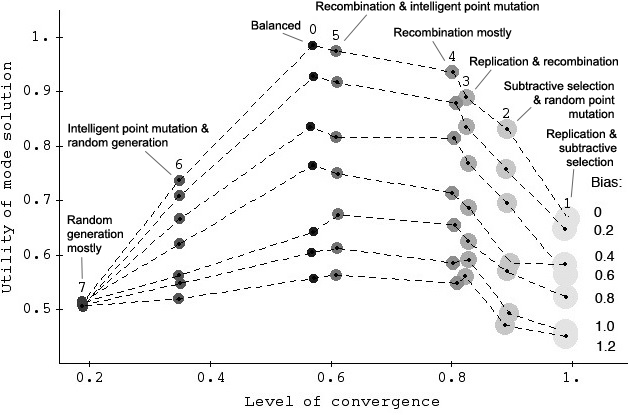}
\end{tabular}
\caption{Computer simulation results (horizontal: level of
  convergence, vertical: true utility value of the most supported
  idea). (a) Effects of within-group heterogeneity $\nu$ and
  group-level bias $\beta$ on two outcome metrics. $\nu=0$ represents
  the case of completely homogeneous groups, while larger values of
  $\nu$ represent more heterogeneous group cases. Larger values of
  bias ($\beta$) represent large discrepancies between the master
  utility function of the group and the true utility function. (b)
  Effects of variations in group-level behavioral patterns on two
  outcome metrics. Within-group heterogeneity $\nu$ was set to 0. See
  text for details of the behavioral patterns used in each of the
  seven different groups.}
\label{comp-model-results}
\end{figure}

Figure \ref{comp-model-results}(a) shows the results of computer
simulations, plotting the group performance in a 2-D space using the
two metrics described above. It was assumed that each agent randomly
chose one of the six evolutionary operators with equal probability in
each iteration. The results clearly showed that both within-group
heterogeneity and group-level bias had negative effects on the true
utility value of the most supported idea. While the group-level bias
had no adverse effect on the level of convergence, the within-group
heterogeneity reduced the level of convergence significantly. The
utility achieved by the most heterogeneous groups dropped to just
above 0.5, meaning that there was no net improvement achieved during
the group discussion. This was due to the intra-group conflicts of
interest among the group members.

We note that the ``balanced'' acts of group members assumed in the
above experiment may be too ideal as a model of actual group members,
because real human groups may have biased behavioral patterns. We
therefore ran another set of experiments using the same simulation
model with different behavioral patterns assumed for different
groups. In forming different group properties, we modeled some
operators singularly (e.g., random generation was the only dominant
operator within the group), and for other groups we combined two
evolutionary operators to reflect increasing complexity of group
behavior (e.g., recombination and intelligent point mutation). For the
former cases, group members were assumed to choose the designated
operator for 95\% of their total actions, with 1\% for each of the
other five operators. For the latter combined cases, they were assumed
to choose each of the two operators for 48\% of their total actions
(96\% in total), with 1\% for each of the other four
operators. Obviously, there are thousands of possible combinations of
operators. For this study, however, we limited our examination to
eight group types: replication and subtractive selection (Group 1);
subtractive selection and random point mutation (Group 2); replication
and recombination (Group 3); recombination mostly (Group 4);
recombination and intelligent point mutation (Group 5); intelligent
point mutation and random generation (Group 6); random generation
mostly (Group 7); and, finally, the balanced team we used in the
previous experiment (Group 0).

Figure \ref{comp-model-results}(b) shows the results. The behaviorally
most diverse Group 0 achieved the highest utility value of the most
supported idea. Meanwhile, a variety of different group performances
were achieved by groups with different balances between
selection-oriented and variation-oriented operators, seen as a
convergence-utility Pareto frontier near the upper-right corner of the
performance space. It is also notable that the random generation
operators (used in Groups 6 and 7) were generally harmful to group
performance.

We derived from these computer simulations the following three
hypotheses: {\em
\begin{description}
\item[\em H1:] Within-group homogeneity of problem understanding improves
  group performance.
\item[\em H2:] Balance between critical and creative attitudes improves
  group performance.
\item[\em H3:] Diversity of available evolutionary operators improves group
  performance.
\end{description}
}

To test these hypotheses experimentally, we designed and conducted
three human subject experiments, which will be described in the
following sections. These experiments were mostly conducted in the
course `BE-461: Exploring Social Dynamics' offered to juniors and
seniors in the Bioengineering and Management programs at Binghamton
University. The study was reviewed and approved by the Binghamton
University IRB. No personally identifiable information was collected
from the subjects.

\section{Human Subject Experiment 1: Product Name Design}
\label{prod}

The first experiment aimed to test Hypothesis H1: Within-group
homogeneity of problem understanding improves group performance. The
experiments were conducted twice in two separate years (2008 and 2011)
using the same protocol described below.

Students were divided into groups of four (or sometimes three). Half
of the groups were made of students of the same gender, in the same
major, and in the same graduation year, which were intended to
represent groups with more homogeneous problem understanding ({\em
  homogeneous condition}). The other half of the groups were made so
that the within-group difference of gender, major and year would be as
high as possible, which were intended to represent groups with more
heterogeneous problem understanding ({\em heterogeneous
  condition}). These groupings were automatically generated by a
computer program, but the conditions were hidden from the students. A
total of 50 students participated in this experiment, forming 13
groups. The following instruction was given to each group leader:
\begin{quote}
{\em You are the leader of a marketing team in a trading company that
  is about to introduce to the U.S.\ market a new cell phone
  manufactured by an Asian country. This product has several key
  features: easy-to-use keypad; easy-to-read display; energy
  efficiency; long battery life; Bluetooth and WiFi ready. The
  original name of this product was in foreign language that doesn't
  sound well to English-speaking people. Your task is to coordinate
  the team discussion and come up with an attractive name of this
  product for U.S.\ customers.  }
\end{quote}
A picture of an ordinary cell phone was also handed to each group to
promote their discussion. Once the group reached a consensus, they
submitted both their final design (name of the phone) and the whole
list of ideas (candidate names) they discussed. Their final designs
were then projected to the screen in the classroom and the students
ranked the final designs individually and independently. This peer
evaluation was used as a measure to quantitatively assess the
utilities of the final designs made by each group. The length of the
list of all the ideas was also measured as a characteristic of the
decision making processes.

The final designs created by the groups under two different conditions
are listed in Supplemental Materials (Table S2). Figure
\ref{prod-results} compares the numbers of ideas generated and the
average ranking scores of the final design between the two
conditions. While the sample size was very small, a statistically
significant difference was detected in terms of the number of ideas
generated (Fig. \ref{prod-results}(a)), i.e., the heterogeneous groups
produced more ideas. This can be understood in that the convergence of
discussion was relatively easier in homogeneous groups so they did not
explore the problem space as much as the heterogeneous groups did. In
the meantime, there was no statistically significant difference
detected regarding the ranking score of final designs between those
two conditions; in fact, groups in the heterogeneous condition
appeared to have produced slightly better names
(Fig. \ref{prod-results}(b)), which may look opposite to what H1 would
imply. However, this result actually makes sense when the possible
effect of group-level bias is considered. As shown in
Fig. \ref{comp-model-results}(a), the utility of final designs
decreases by not only within-group heterogeneity but also group-level
bias. It is reasonable to assume that the more diverse the group
members are, the better they represent the true utility function
(i.e., the preference of the whole class, in this case). Therefore, we
interpret this result as a mixture of two different effects of
within-group heterogeneity on the quality of final designs---to
decrease it due to intra-group conflicts, and to increase it by
reducing potential group-level biases.

\begin{figure}[t]
\centering
\begin{tabular}{ll}
(a) & (b) \\
\includegraphics[width=0.3\columnwidth]{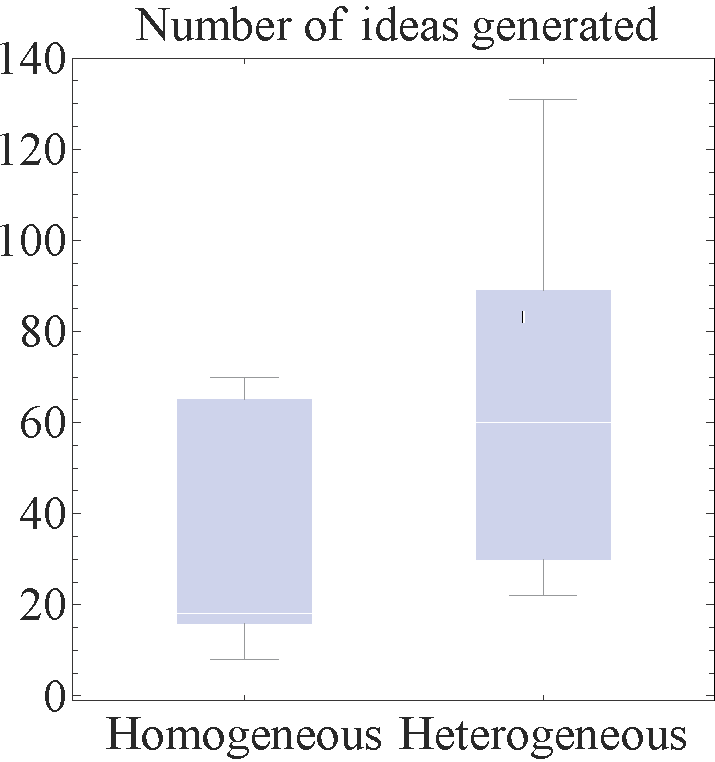} &
\includegraphics[width=0.3\columnwidth]{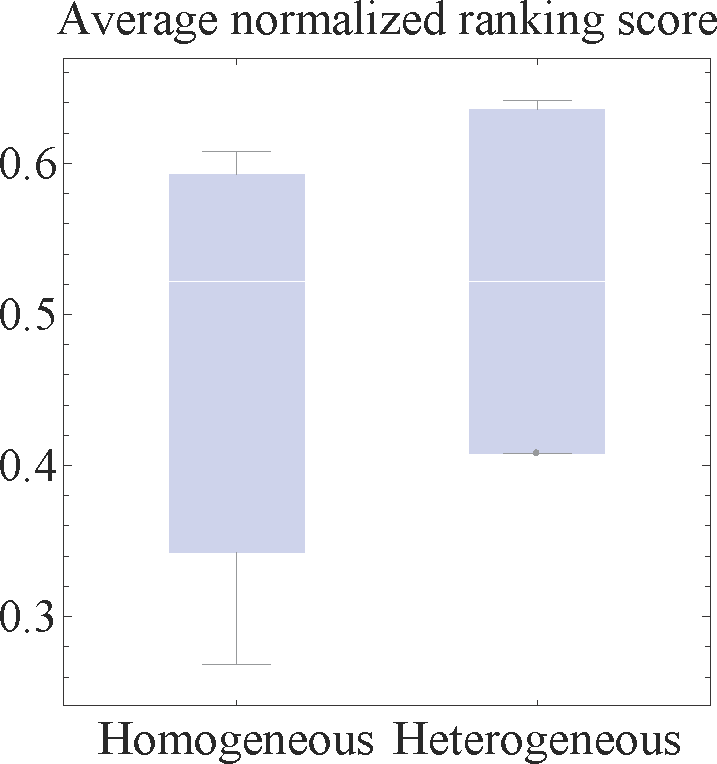} \\
\end{tabular}
\caption{Results of Experiment 1, comparing the distributions of
  outcomes between homogeneous and heterogeneous groups. (a) Number of
  ideas generated. A statistically significant difference was detected
  between the conditions ($p= 0.0456$). (b) Average normalized ranking
  score of final designs (1 = best, 0 = worst). No statistically
  significant difference was detected ($p=0.2816$).}
\label{prod-results}
\end{figure}

\section{Human Subject Experiment 2: Catch Phrase Design}
\label{catch}

The second experiment aimed to test Hypothesis H2: Balance between
critical and creative attitudes improves group performance. The
experiments were conducted twice in two separate years (2008 and 2010)
using the same protocol described below.

Students were randomly divided into groups of four (or sometimes
three). A total of 45 students participated in this experiment,
forming 12 groups. The following instruction was given to each group
leader:
\begin{quote}
{\em 
You are the leader of a marketing team in a manufacturer of a new
laptop computer. The specs of the new laptop are quite mediocre, with
no technical feature truly novel and attractive to customers. In any
case, your task is to coordinate the team discussion and come up with
a list of inspiring catch phrases to promote the sales of the new
laptop.
}
\end{quote}
A picture of an ordinary laptop was also handed to each group to
promote their discussion. Three different experimental conditions were
created by providing the following additional information to selected
groups before the discussion:
\begin{quote}
{\em {\bf Critical condition:} {\rm (4 groups)} When coordinating
  discussion, share the following principles with your team members:
  Promote and maintain critical attitude throughout the
  discussion. Always play devil's advocate, trying to find ways for
  each catch phrase to be potentially problematic.  Incremental
  improvement of existing ideas is the key to making a reliable
  idea. Completely new ideas will never be better than well-tested
  ideas.}
\end{quote}

\begin{quote}
{\em {\bf Creative condition:} {\rm (4 groups)} When coordinating
  discussion, share the following principles with your team members:
  Promote and maintain creative attitude throughout the
  discussion. Always give positive feedback to someone who presented a
  new idea, trying to find good aspects in it.  Crazy inspiration and
  idiosyncratic thinking is the key to breaking the barrier of
  stereotyped ideas. Incremental improvement of existing ideas will
  never work out.}
\end{quote}

\begin{quote}
{\bf Control condition:} (4 groups) No additional instruction was
given.
\end{quote}

As given in the instruction above, the groups were initially asked to
simply produce a list of catch phrases, but after 20 minutes of
discussion, they were told to make a final design and choose the best
catch phrase out of the produced list. Once the team reached a
consensus, they submitted both their final design (catch phrase) and
the whole list of ideas (candidate phrases) they discussed. The length
of the list of all the ideas was also measured as a characteristic of
the decision making processes.

Unlike in the first experiment, the final designs were evaluated by a
large number of third parties, i.e., students in another undergraduate
course on organizational behavior that had more than 120
enrollments. All the final designs were projected in a randomized
order on a screen of a large lecture hall, and then those students in
the organizational behavior course rated the quality of each product
individually and independently. The average of those rating results
was used as the average rating score of each design.

\begin{figure}[t]
\centering
\begin{tabular}{ll}
(a) & (b)\\
\includegraphics[width=0.3\columnwidth]{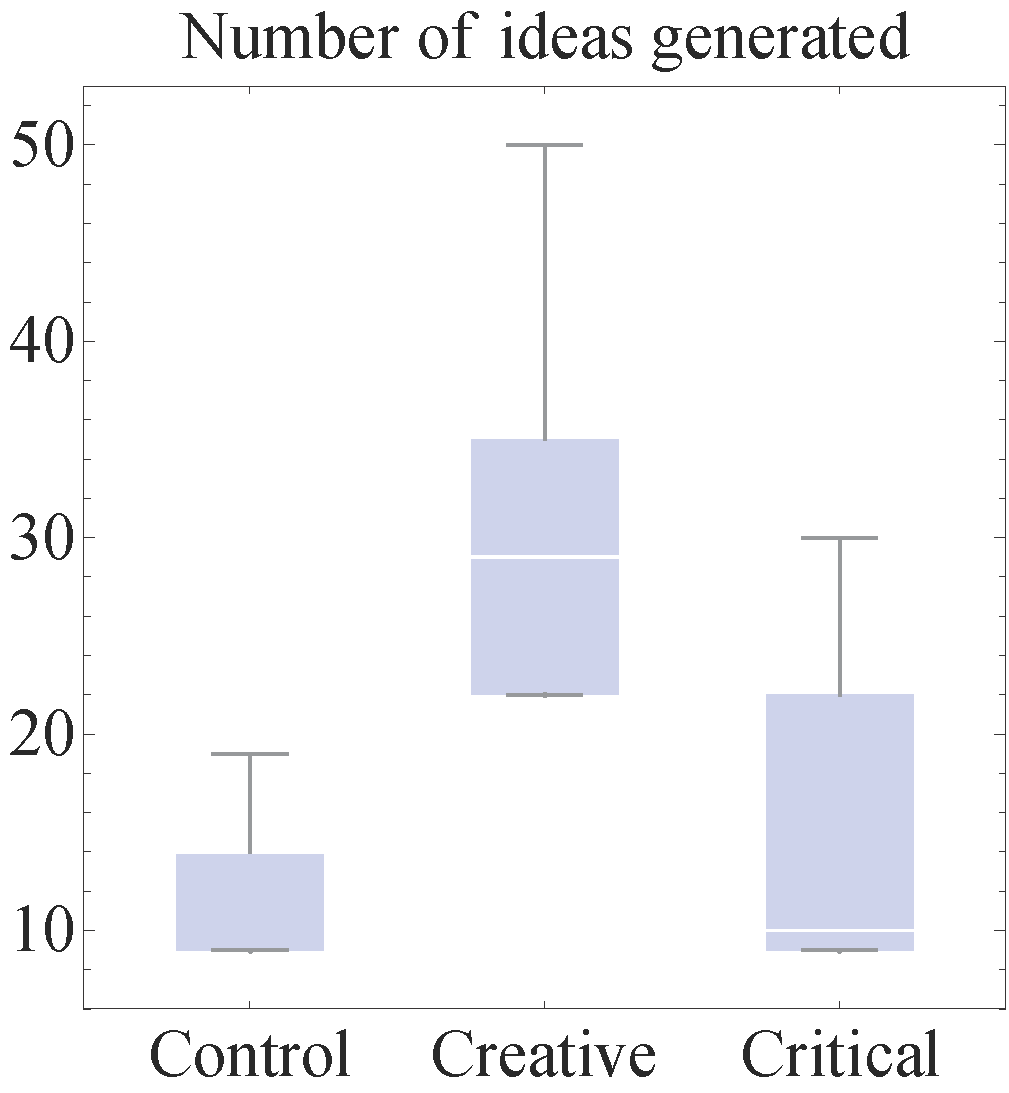} &
\includegraphics[width=0.32\columnwidth]{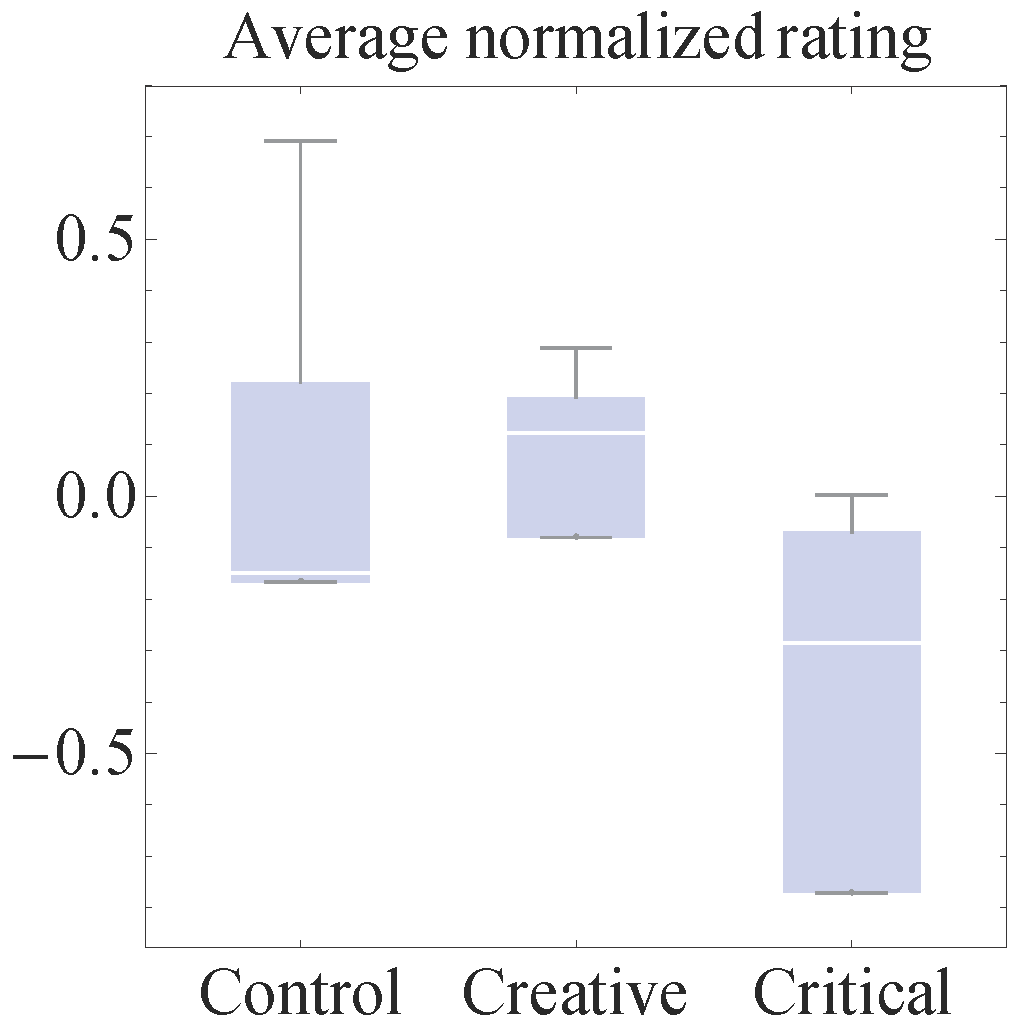}
\end{tabular}
\caption{Results of Experiment 2, comparing the mean outcomes between
  control, creative and critical groups. (a) Number of ideas
  generated. A statistically significant difference was detected among
  the three conditions by ANOVA ($p=0.0314$), and Tukey's and
  Bonferroni's posthoc tests showed that there was a significant
  difference between control and creative conditions. (b) Average
  normalized rating score of final designs (positive: good, negative:
  bad). No statistically significant difference was detected by
  ANOVA ($p=0.156$).}
\label{catch-results}
\end{figure}

The final designs created by the groups under three different
conditions are listed in Supplemental Materials (Table S3). Figure
\ref{catch-results} compares the numbers of ideas generated and the
average rating scores of the final design between the three
conditions. Again, the sample size was quite small for this
experiment, but a statistically significant difference was detected
among those conditions in terms of the number of ideas
generated. Specifically, creative groups produced significantly more
ideas than control groups (and seemingly more than critical groups,
though with no statistical significance). Also, it was indicated that
designs made by critical groups may have received lower ratings than
those made by others.

\begin{figure}[t]
\centering
\begin{tabular}{l}
(a) Sample from creative condition \\
\includegraphics[width=\columnwidth]{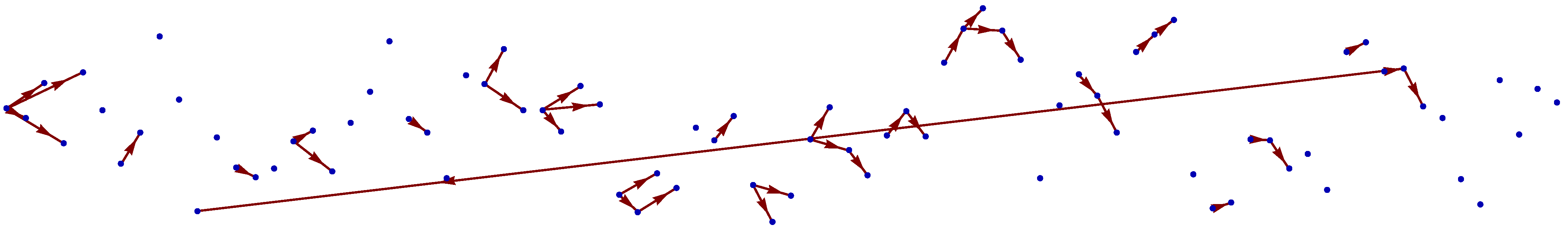}\\
~\\
(b) Sample from critical condition\\
\multicolumn{1}{c}{\includegraphics[width=0.8\columnwidth]{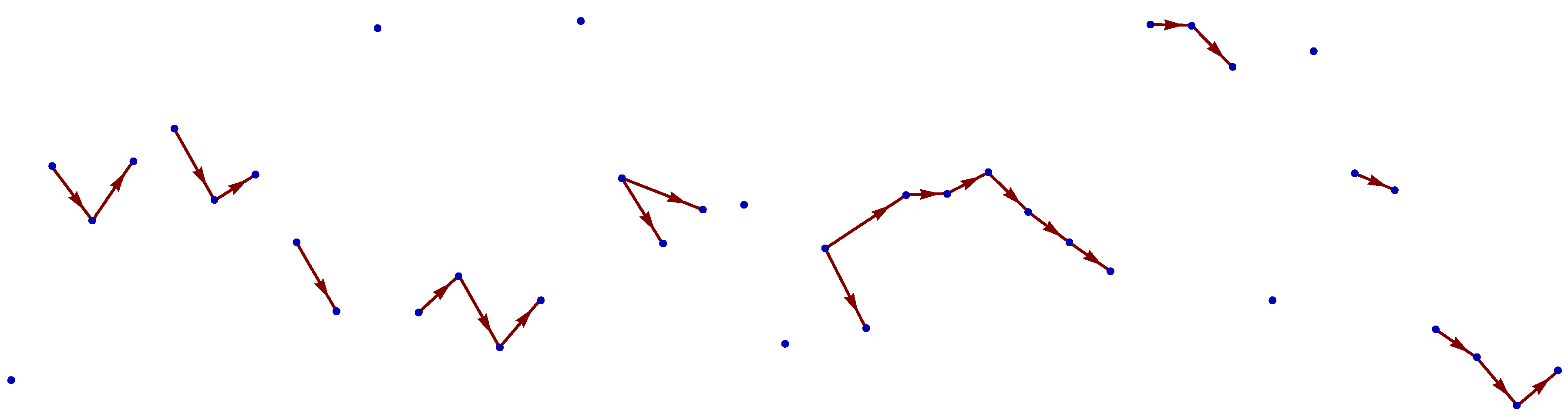}}\\
\end{tabular}
\caption{Sample idea genealogies constructed from transcripts of
  recordings of discussions in Experiment 2. The horizontal axis
  represents time (not scaled to real time, but arranged in the order
  of ideas' first appearance in the discussion). The vertical axis was
  arbitrarily set just for visualization purposes. Each small dot
  represents a new idea (a catch phrase). An arrow connecting two
  ideas indicates that a new idea came out of an existing idea, which
  was detected and noted by human transcribers.}
\label{genealogy-manual}
\end{figure}

We also transcribed a few sample recordings of the discussions taken
in this experiment and manually constructed genealogies of ideas that
show when new ideas emerged from which existing ideas. Figure
\ref{genealogy-manual} shows two illustrative cases, one from a
creative group and the other from a critical group. The creative group
not only produced more ideas than its critical counterpart, but also
showed more ``branching'' ideation processes to explore various
possibilities (Fig. \ref{genealogy-manual}(a)). In contrast, the
critical group's idea genealogy was mostly sequential with very few
branching events (Fig. \ref{genealogy-manual}(b)), implying that they
produced catch phrases through incremental revisions.

An interesting observation in these results is that the control groups
seem to have made collective decisions most efficiently---they
conducted as little evolutionary exploration as the critical groups
(Fig. \ref{catch-results}(a)) and still produced quality solutions
comparable to those from the creative groups
(Fig. \ref{catch-results}(b)). In other words, collective human
decision making apparently worked optimally when no additional
instruction was given. We interpret these results as follows:
Behaviors of people working in groups are most balanced when no
explicit instructions are given, leading to best decision
outcomes. Instructing teams to be either creative or critical may
result in loss of behavioral balance and therefore less efficient or
less productive discussion, as already seen in our computer
simulations (Fig. \ref{comp-model-results}(b)). These findings and
observations support Hypothesis H2, that the balance between critical
and creative attitudes improves group performance.

\section{Human Subject Experiment 3: Collaborative Design with Hyperinteractive Evolutionary Computation}

\label{swarm}

The third experiment aimed to test Hypothesis H3: Diversity of
available evolutionary operators improves group performance. This
naturally follows the results of the second experiment discussed
above. For the third experiment, we used interactive evolutionary
computation (IEC) \cite{iec} as an experimental platform, and
developed a new framework called ``Hyperinteractive Evolutionary
Computation (HIEC)'' that offers users more controllability than
conventional IEC. More details of HIEC can be found in \cite{hiec},
and its overview is also provided in Supplemental Materials. This
experiment consisted of two parts: (i) directly manipulating the
availability of evolutionary operators and (ii) electronically
recording their usage.

The first part of the experiment took place in 2008. A total of 21
students were randomly divided into 7 groups. Each group was given 10
minutes to collaboratively design an ``interesting'' swarming pattern
produced using the Swarm Chemistry model
\cite{swarmchemistry,ieeealife2009}. The following four conditions
were configured in the HIEC software and assigned randomly to each
group:
\begin{itemize}
\item {\em Baseline condition:} Neither mixing nor mutation operators
  were available.
\item {\em Mixing condition:} Only the operator for mixing two
  patterns was available.
\item {\em Mutation condition:} Only the operator for mutating a
  pattern was available.
\item {\em Mixing + mutation condition:} Both the mixing and mutation
  operators were available.
\end{itemize}
The design process was repeated three times (each time group members
were randomly shuffled) so that $3 \times 7 = 21$ final designs were
produced. Those final designs were projected to the screen in the
classroom in a randomized order, and then the students evaluated how
``cool'' each swarm was on a 10-point scale individually and
independently. As a result, each design received 21 individual rating
scores. The peer evaluation was used to quantitatively assess the
quality of the final designs made in each condition.

\begin{SCfigure}
\centering
\includegraphics[width=0.5\columnwidth]{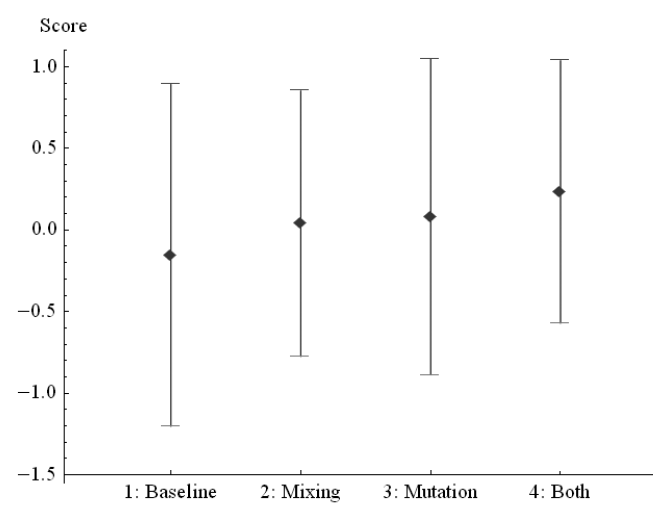}
\caption{Results of Experiment 3(i). Comparison of average normalized
  rating scores of final designs produced by groups under different
  conditions. Mean normalized scores are shown by diamonds, with error
  bars around them showing standard deviations. A statistically
  significant difference was detected among the four conditions by
  ANOVA ($p=0.0151$), while Tukey's and Bonferroni's post-hoc tests
  detected a significant difference between conditions 1 and 4.}
\label{means}
\end{SCfigure}

Students' evaluation results were first normalized so that the mean
was 0 and the standard deviation 1 for each individual student's
responses, in order to equalize the contribution of each student's
ratings to the overall statistics. Then the normalized scores were
collected and averaged for each experimental condition. Figure
\ref{means} summarizes the results. A statistically significant
difference was detected among different conditions. Tukey's and
Bonferroni's post-hoc tests detected a significant difference between
conditions 1 and 4, supporting Hypothesis H3, i.e., the availability
of more evolutionary operators makes the outcomes of group decision
making better. It was also observed that each of the Mixing and
Mutation operators contributed nearly independently to the improvement
of the design quality \cite{ieeealife2009}.

The second part of the experiment used a modified version of the HIEC
software so that we could electronically record a complete
time-stamped log of every single evolutionary event that happened in
the same swarm design task. Such a detailed log of decision making
processes allowed us to quantitatively analyze how variation and
selection occurred in human decision making. This part of the
experiment was conducted twice in two separate years (in 2009 and
2010) using the same protocol described below.

Students were randomly divided into groups of three (or sometimes
four). Each group was assigned to a station with a touch-screen PC
running the modified HIEC software. Each group was given 10 minutes to
work on the same collaborative design task as in the first part, with
no further guidance given. This phase of the experiment served as the
experimental control ({\em Control condition}). Then the subjects were
reshuffled into new groups. Half of the groups were primed to be
critical and risk-averse, with the following written instruction:
\begin{quote}
{\em 
{\bf Critical condition: }
Promote and maintain critical attitude throughout the design
process. Incremental improvement of existing designs is the key to
making a reliable idea. Completely new designs will never be better
than well-tested ones.
}
\end{quote}
The other half of them were primed to be creative and adventurous,
with the following written instruction:
\begin{quote}
{\em
{\bf Creative condition:}
Promote and maintain creative attitude throughout the design
process. Crazy inspiration and idiosyncratic thinking is the key to
breaking the barrier of stereotyped designs. Incremental improvement
of existing designs will never work out.
}
\end{quote}
Then the groups were once again given 10 minutes to work on the
collaborative design task. Finally, the above step was repeated one
more time after another group reshuffling.

A total of 40 students participated in this experiment, forming 39
sessions (= 13 groups x 3 times). Three sessions had technical
problems during the experiment and thus their data were excluded from
the analysis. As a result, we collected data from 11 sessions working
under the control condition, 12 under the creative condition, and 13
under the critical condition. A log file containing detailed
information about all the evolutionary events in each session was
saved in a local hard drive of each PC, and later collected for
post-experimental analysis.

Similarly to Experiment 2, the final designs were evaluated by third
parties, i.e., 22 students who took the same course BE-461 in the
following year. All the final designs were projected in a randomized
order on a screen in the classroom, and then those students rated the
quality of each product on a 0-10 scale individually and
independently. The average of those rating results was used as the
average rating score of each design.

\begin{figure}[t]
\centering
\begin{tabular}{lll}
(a) & (b) & (c)\\
\includegraphics[width=0.3\columnwidth]{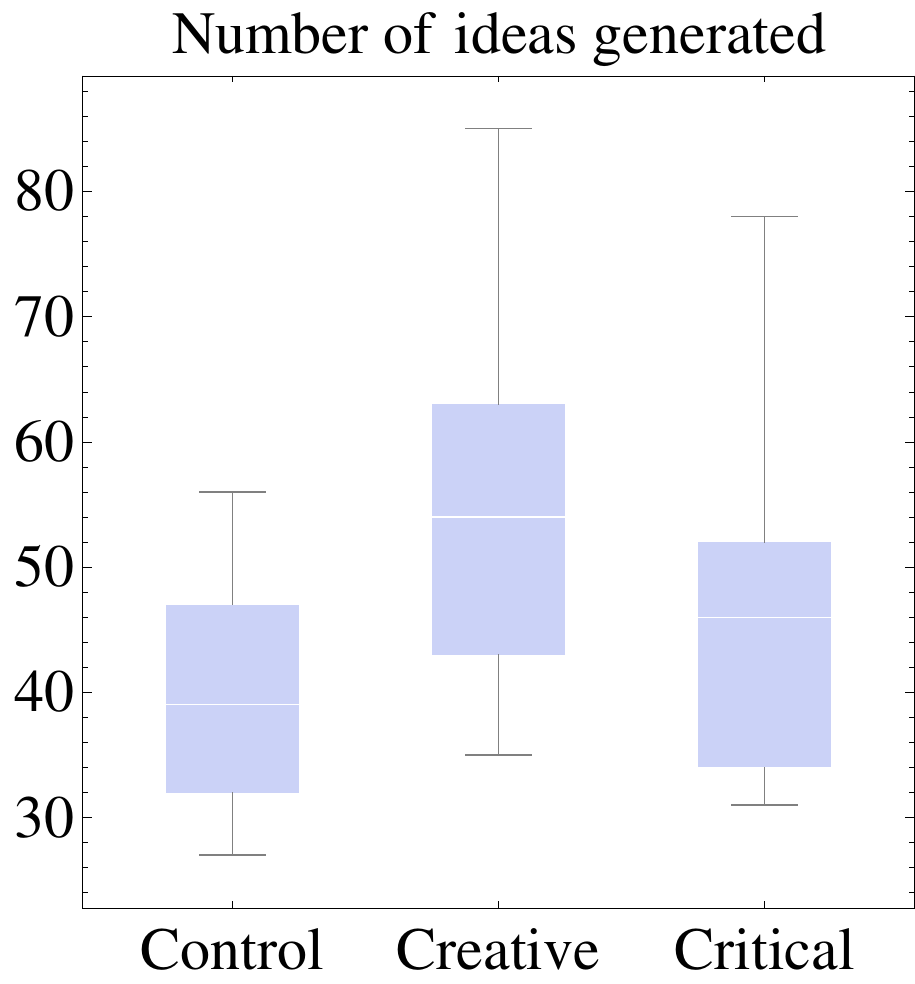} &
\includegraphics[width=0.32\columnwidth]{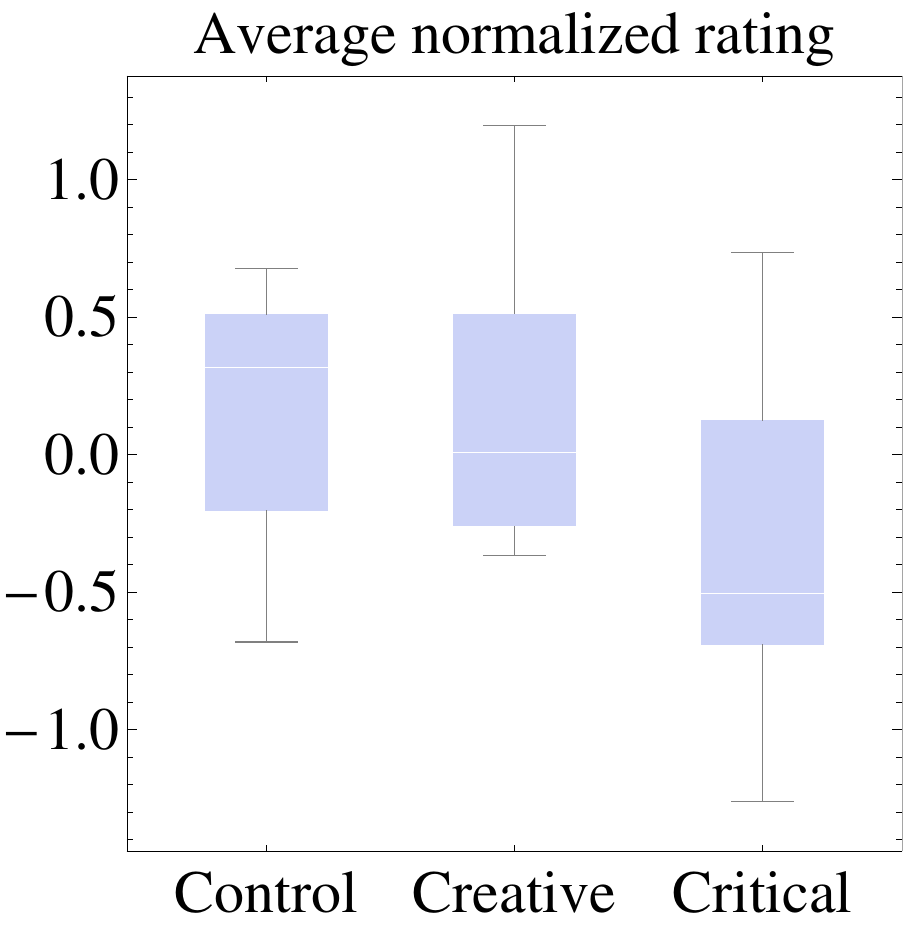} &
\includegraphics[width=0.3\columnwidth]{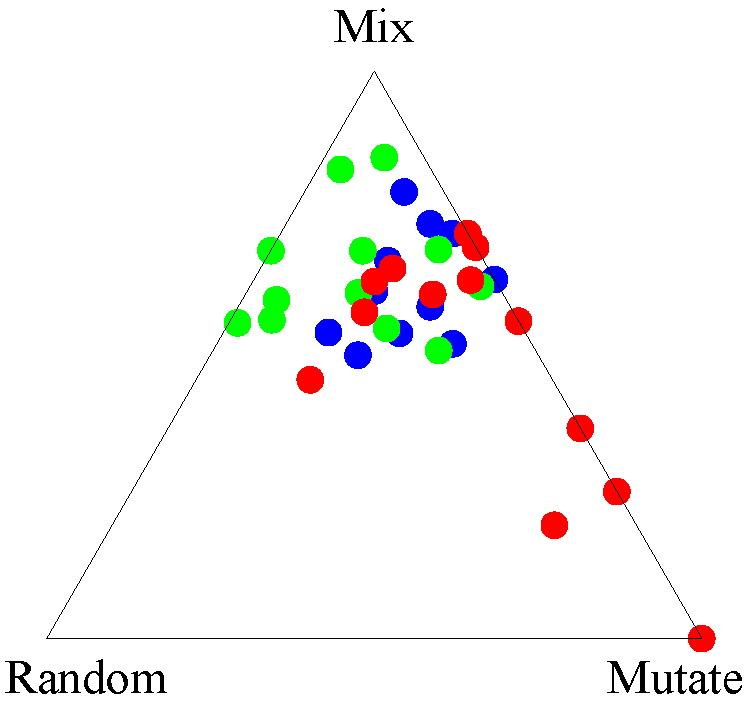}
\end{tabular}
\caption{Results of Experiment 3(ii), comparing the distributions of
  outcomes between three conditions. (a) Number of ideas generated. A
  moderate difference was detected among the three conditions by ANOVA
  ($p=0.0547$), while Tukey's posthoc test indicated a significant
  difference between control and creative conditions. (b) Average
  normalized rating score of final designs (positive: good, negative:
  bad). No statistically significant difference was detected by ANOVA
  ($p=0.0797$). (c) Ternary plot of the group behavior data with
  respect to the mix, mutate, and random evolutionary operators. Each
  marker represents data taken from one group (blue: control, green:
  creative, red: critical).}
\label{swarm-results}
\end{figure}

Figure \ref{swarm-results} summarizes the results, which were
strikingly similar to those of the previous experiment
(Fig. \ref{catch-results}); the creative groups produced most ideas in
this experiment (Fig. \ref{swarm-results}(a)) while the quality of
final designs produced by control groups were comparable to those by
creative groups (Fig. \ref{swarm-results}(b)). It was also observed
that creative groups tended to use more mixing operators, critical
groups focused more on mutation, and control groups sat somewhere in
between (Fig. \ref{swarm-results}(c)), which supports the observation
and interpretation obtained in Experiment 2. These findings clearly
show that priming conditions did affect groups' attitudes in
discussion, directly relating human behavior in decision making with
evolutionary concepts.

Moreover, the use of HIEC software made it possible to collect
detailed trajectories of how ideas have been explored and how a final
decision has been developed over time. Figure S2 in
Supplemental Materials presents visualization of a sample genealogy of
ideas generated based on the actual data taken from one of the groups
in this experiment. Visual inspection of the idea genealogies indicate
that there are some topological variations associated with the priming
conditions. More systematic, quantitative analysis of topologies of
those genealogies is to be reported elsewhere.

\section{Conclusions}

We have presented a series of experiments that illustrate how the
dynamics of collective human decision making processes can be studied
as evolutionary processes. The results obtained from our human subject
experiments nicely matched the predictions made by the computational
experiments, illustrating the validity of evolutionary understanding
of human decision making and creativity processes. In those
experiments, evolutionary computation (EC) played multiple roles, as a
theoretical/computational modeling framework as well as an
experimental tool for data collection. We believe our work
demonstrates untapped potential of EC for interdisciplinary research
involving human and social dynamics.

One particularly important take-home message this project has provided
us with is the nontrivial role of ``diversity'' in groups. While the
diversity of problem understanding may cause intra-group conflicts and
thereby harm the group performance, the diversity of behavior in
discussion (e.g., balance between variation and selection) can offer
various evolutionary paths in decision making processes that will help
improve the group performance. Our results also led us to a conjecture
that humans are naturally most balanced in their behavior, which could
also be explained from an evolutionary viewpoint.

The key findings obtained in this project support our framework that
uses evolutionary principles to describe collective human decision
making and creativity. This research has presented a conceptual as
well as technical shift of focus from human individuals to the ideas
evolving through discussions. We hope that this will lead to a
theoretical advancement from a traditional, individually-focused
psychological or social science paradigm to a more dynamic,
multilevel, evolutionary paradigm for collective social processes.

For more information about this project, see the official project
website \cite{website}.

\section*{Acknowledgments}

We thank Francis J. Yammarino, Craig Laramee, David Sloan Wilson,
J. David Schaffer, Dene Farrell, Andrew Talia, Benjamin James Bush,
Chanyu Hao, Hadassah Head, Thomas Raway, Andra Serban, Alka Gupta and
Jeffrey Schmidt for their help in conducting this project. This work
was supported in part by the National Science Foundation, under Grants
SES-0826711 and DUE-0737313.

\end{document}